\documentclass[a4paper, 10pt, conference]{IEEEtran}      % Use this line for a4
\usepackage{FG2018}

\FGfinalcopy % *** Uncomment this line for the final submission

\usepackage{graphics} % for pdf, bitmapped graphics files
\usepackage{epsfig} % for postscript graphics files
\usepackage{amsmath} % assumes amsmath package installed
\usepackage{amssymb}  % assumes amsmath package installed
\usepackage{amsfonts}
\usepackage{multirow}
\usepackage{stfloats}

\def\FGPaperID{****} % *** Enter the FG 2018 Paper ID here

\title{\LARGE \bf
Toward Marker-free 3D Pose Estimation in Lifting:\\ 
A Deep Multi-view Solution
}

\author{\IEEEauthorblockN{Rahil Mehrizi\IEEEauthorrefmark{1},
Xi Peng\IEEEauthorrefmark{2},
Zhiqiang Tang\IEEEauthorrefmark{2}, 
Xu Xu\IEEEauthorrefmark{5},
Dimitris Metaxas\IEEEauthorrefmark{2} and
Kang Li\IEEEauthorrefmark{1}\IEEEauthorrefmark{2}\IEEEauthorrefmark{4}}\\
\IEEEauthorblockA{\IEEEauthorrefmark{1}Department of Industrial \& Systems Engineering\\
Rutgers University,
Piscataway, New Jersey, USA\\ Email: rahil.mehrizi@rutgers.edu}
\IEEEauthorblockA{\IEEEauthorrefmark{2}Department of Computer Science, Rutgers University, Piscataway, New Jersey, USA}
\IEEEauthorblockA{\IEEEauthorrefmark{4}Department of Orthopaedics, Rutgers New Jersey Medical School, Newark, New Jersey, USA}
\IEEEauthorblockA{\IEEEauthorrefmark{5}Department of Industrial and Systems Engineering, North Carolina State University, Raleigh, NC, USA}}

\begin{document}

\ifFGfinal
\thispagestyle{empty}
\pagestyle{empty}
\else
\author{Anonymous FG 2018 submission\\ Paper ID \FGPaperID \\}
\pagestyle{plain}
\fi
\maketitle

%%%%%%%%%%%%%%%%%%%%%%%%%%%%%%%%%%%%%%%%%%%%%%%%%%%%%%%%%%%%%%%%%%%%%%%%%%%%%%%%
\begin{abstract}
Lifting is a common manual material handling task performed in the workplaces. It is considered as one of the main risk factors for Work-related Musculoskeletal Disorders. To improve work place safety, it is necessary to assess musculoskeletal and biomechanical risk exposures associated with these tasks, which requires very accurate 3D pose. Existing approaches mainly utilize marker-based sensors to collect 3D information. However, these methods are usually expensive to setup, time-consuming in process, and sensitive to the surrounding environment. In this study, we propose a multi-view based deep perceptron approach to address aforementioned limitations. Our approach consists of two modules: a "view-specific perceptron" network extracts rich information independently from the image of view, which includes both 2D shape and hierarchical texture information; while a "multi-view integration" network synthesizes information from all available views to predict accurate 3D pose. To fully evaluate our approach, we carried out comprehensive experiments to compare different variants of our design. The results prove that our approach achieves comparable performance with former marker-based methods, i.e. an average error of $14.72 \pm 2.96$ mm on the lifting dataset. The results are also compared with state-of-the-art methods on HumanEva-I dataset \cite{sigal2010humaneva}, which demonstrates the superior performance of our approach.
\end{abstract}


%%%%%%%%%%%%%%%%%%%%%%%%%%%%%%%%%%%%%%%%%%%%%%%%%%%%%%%%%%%%%%%%%%%%%%%%%%%%%%%%
\section{INTRODUCTION}
Work-related musculoskeletal disorders (WMSD) are commonly observed among the workers involved in material handling tasks such as lifting. To improve work place safety and decrease the risk of WMSD, it is necessary to analyze biomechanical risk exposures associated with these tasks by capturing the body pose and assessing joints kinematic and critical joints stress. In recent years, several systems were developed to capture the 3D body pose and assess the movement of workers, which roughly can be categorized into two groups: direct measurement \cite{david2005ergonomic} and observational systems \cite{spielholz2001comparison}.
 
Direct measurement systems require motion capture equipment and attachment of the reflective markers on the subject's body to capture the 3D coordination of the body joints. They are considered as a relatively reliable and accurate system for estimation of the joints kinematics. However, the wide spread use of direct measurement systems are limited due to its limitations. First, they require expensive motion capture equipment; second, attaching markers to the subject's body is time consuming and can obstruct the subject's activities.

Observational systems like video-based coding system, on the other hand, use recorded videos of the subject and extract a few key frames from them. Then, raters estimate the body pose by making an optimal fit of a predefined digital manikin to the selected video frames. Finally, using the estimated body pose data and time information extracted from the videos \cite{peng2016track}, joints trajectory is generated for the entire task by applying a motion pattern prediction algorithm \cite{hsiang1998video}. Observational systems are not as accurate as direct measurement systems and the result accuracy rely on the experience of the rater, especially when joints angle become close to the posture boundaries \cite{coenen2013inter}.

   \begin{figure}[t]
      \centering
      \includegraphics[width=3in]{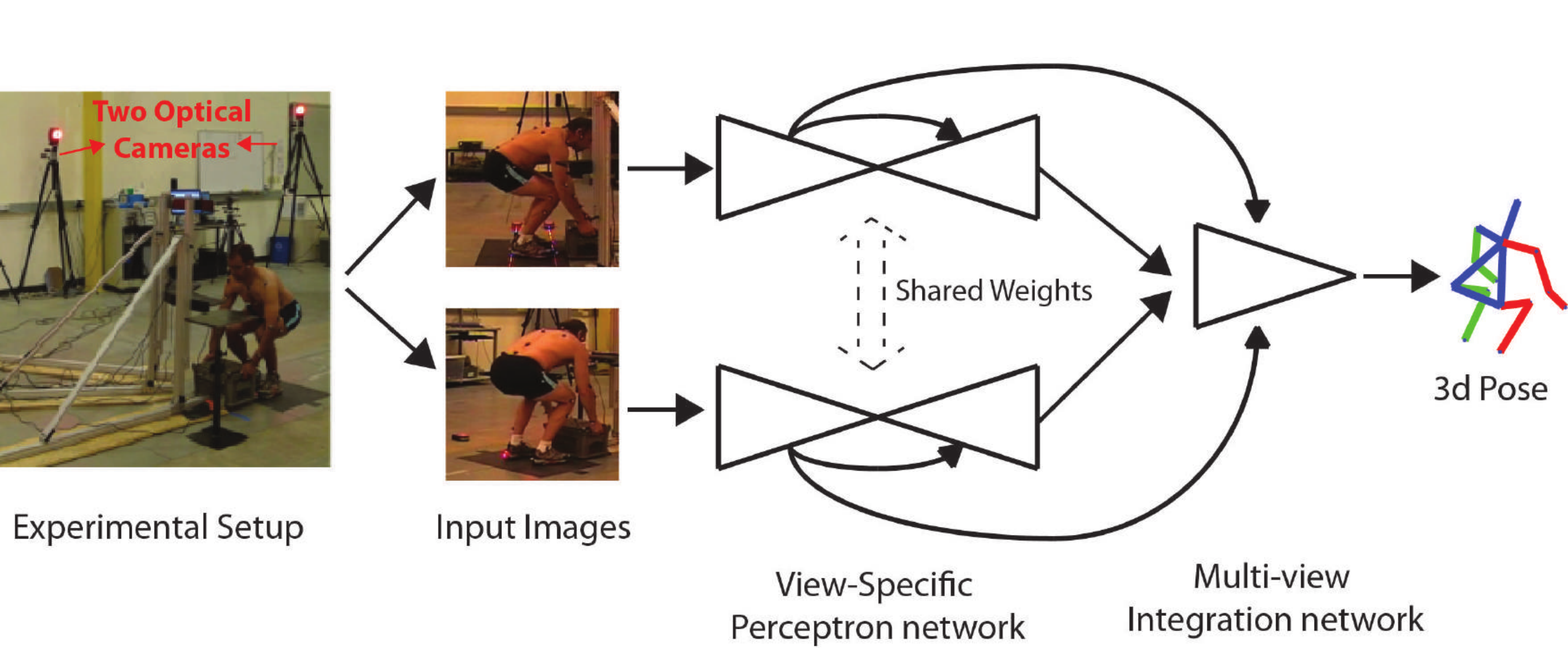}
      \caption{An overview of our approach. The "view-specific perceptron" network extracts both shape and hierarchical texture from different views; while the "multi-view integration" network synthesizes information from all available views to estimation 3D pose. Hierarchical skip connections are not only shared locally inside the first network, but also shared globally between two networks for efficient and effective feature embedding.}
      \label{figurelabel}
   \end{figure}
   
\noindent In this study, we propose a Deep Neural Network (DNN) based multi-view perceptron framework for maker-free 3D motion capture. Fig. 1 illustrates the experimental setup along with the overview of our approach. Our method consists of two networks: a "view-specific perceptron" network extracts both 2D shape and hierarchical texture information from different views \cite{peng2015circle}; while a "multi-view integration" network synthesizes information from all available views to provide accurate 3D pose. It will be shown that sharing hierarchical texture information globally between the two networks in addition to a locally use inside the first network can significantly improve the accuracy. It makes the method suitable for biomechanical analysis, which requires higher accuracy compare to other applications. Since our proposed method eliminates the need of attaching markers onto the subject's body segments or hiring raters to estimate the pose, it can overcome the limitations of both direct measurement and observational systems. To summarize, our contributions are:

\begin{itemize}
  \item We propose a novel DNN-based method to estimate accurate 3D pose from multiple 2D views for biomechanical analysis.
  \item We propose hierarchical skip connections to share rich texture information in different scales, which is proved to 
be crucial for efficient and accurate 3D inference.  
  \item Comprehensive experiments are performed to evaluate different variants of our design, which proves the superior performance of our approach in various aspects.
  \item The results on a real-world multi-view lifting dataset prove that our approach can meet the high-level accuracy requirement in workplace biomechanical analysis.
\end{itemize}
%%%%%%%%%%%%%%%%%%%%%%%%%%%%%%%%%%%%%%%%%%%%%%%%%%%%%%%%%%%%%%%%%%%%%%%%%%%%%%%%
\section{BACKGROUND}
We review related works in two categories. The first category is an overview of the previous work in marker-less posture estimation for biomechanical application, which is important since the focus of this study is on this specific application. The second category, is a summary of recent human pose estimation methods using deep learning.

\subsection{Pose Estimation for Biomechanical Application}
Human pose estimation is important for biomechanical analysis and preventing WMSD. Even though, marker-less pose estimation methods are considered as a potential substitute for the traditional marker-based method, they are not widely studied for biomechanical and clinical applications, which require higher accuracy and robustness in comparison with the other applications \cite{mundermann2006evolution}. There are few studies which explored the field of computer vision and proposed marker-less methods for biomechanical and clinical applications. In particular, \cite{mehrizi2017using, mehrizi2018computer} proposed a computer vision based method for estimation of 3D pose estimation and lower back loads in the symmetrical lifting tasks. In another study by \cite{corazza2010markerless}, a Levenberg-Marquardt minimization scheme over an iterative closest point algorithm was employed to estimate human motion through a marker-less motion capture system. Goffredo et. al. \cite{goffredo2009markerless} proposed a marker-less framework to estimate human pose for a sit-to-stand task by means of a maximum likelihood approach carried out in the Gauss–Laguerre transform domain. These studies demonstrate the feasibility of computer vision approaches for the biomechanical analysis. However, they are limited to a few types of motions and lifting as one of the most common motions in the workplaces and as an important risk factor for WMSD is not fully studied. Additionally, deep learning, which is considered as the state-of-the-art approach in the domain of the vision tasks is not studied for the field of biomechanical application. In this study, we propose a deep learning based framework for maker-less 3D motion capture. It will be shown that using the proposed framework can achieve very high accuracy, which is suitable for the biomechanical analysis.

\subsection{Deep Learning for Human Pose Estimation} 
Earlier computer vision based approaches for 3D human pose estimation used a discriminative or generative method to learn a mapping from the image features to the 3D human pose. All of these approaches suffer from the fact that they utilize hand crafted image features e.g. HOG \cite{dalal2005histograms}, SIFT \cite{muller2010human}, etc. Approaches based on the hand crafted image features are not able to handle heterogeneous or complex datasets \cite{antipov2015learned, sargano2017comprehensive}. With the emergence and advances of deep learning techniques, approaches that employ deep convolutional neural networks to learn the image features \cite{li2017deeprebirth}, have become the standard in the domain of the vision tasks. DNN approaches have achieved the highest performance for several vision tasks such as visual recognition \cite{peng2017reconstruction, di2017large}, image generation \cite{zhang2017generative, zhang2017image}, and human pose estimation \cite{newell2016stacked, wei2016convolutional}. \\
More recent DNN approaches for 3D human pose estimation tend to learn an end-to-end DNN to regress directly from the images to the 3D joints coordination \cite{li20143d, tekin2016direct, rogez2016mocap, peng2016recurrent}. Other DNN approaches, on the other hand, have studied frameworks that employ 2D pose estimation as an intermediate step and leverage this information to infer the 3D pose from it. Chen et. al. \cite{chen20163d} suggests that 2D pose is a useful intermediate representation and can aid the 3D pose estimation. While \cite{chen20163d, park20163d, wang2014robust} represents intermediate 2D pose as 2D coordination of the joints, \cite{pavlakos2017coarse, tome2017lifting, tekin2017learning} define it by a set of heatmaps that encode the probability of observing a specific joint at the corresponding image location. Tome et. al. \cite{tome2017lifting} proposes multi-stage DNN architecture combined with a probabilistic knowledge of 3D human pose, which estimates 2D joints heatmap and 3D pose simultaneously to improve both tasks. Pavlakos et. al. \cite{pavlakos2017coarse} trains a DNN with 2D joints heatmaps as an intermediate representation to predict per voxel likelihood for each joint in the 3D space instead of directly regressing the 3D joints coordination. They uses a coarse-to-fine technique to overcome the high dimensionality problem of the volumetric representation. They also suggests combining 2D joints heatmaps with image features for the intermediate representation, to take advantage of both the image cues along with the reliably detected heatmaps. The same combination is applied in a study by \cite{tekin2017learning} in which 2D joints heatmaps and images are fed into a two-stream architecture, Then the combination of these two streams are then fed into a fusion stream at a specific layer to obtain the final 3D human pose estimate.\\
In this study, we propose a novel deep learning based method to estimate 3D human pose from multi-view images. Instead of using the 2D heatmap as the only intermediate supervision, we propose to share both shape and hierarchical texture locally and globally for efficient 3D inference. In contrast to the recent work in 3D human pose estimation whose focus are on single view and challenging setting, the proposed network is designed to handle multi-view images, which is a common setup for the biomechanical analysis experiments.

%%%%%%%%%%%%%%%%%%%%%%%%%%%%%%%%%%%%%%%%%%%%%%%%%%%%%%%%%%%%%%%%%%%%%%%%%%%%%%%%
  \begin{figure*}[thpb]
      \centering
      \includegraphics[width=5in]{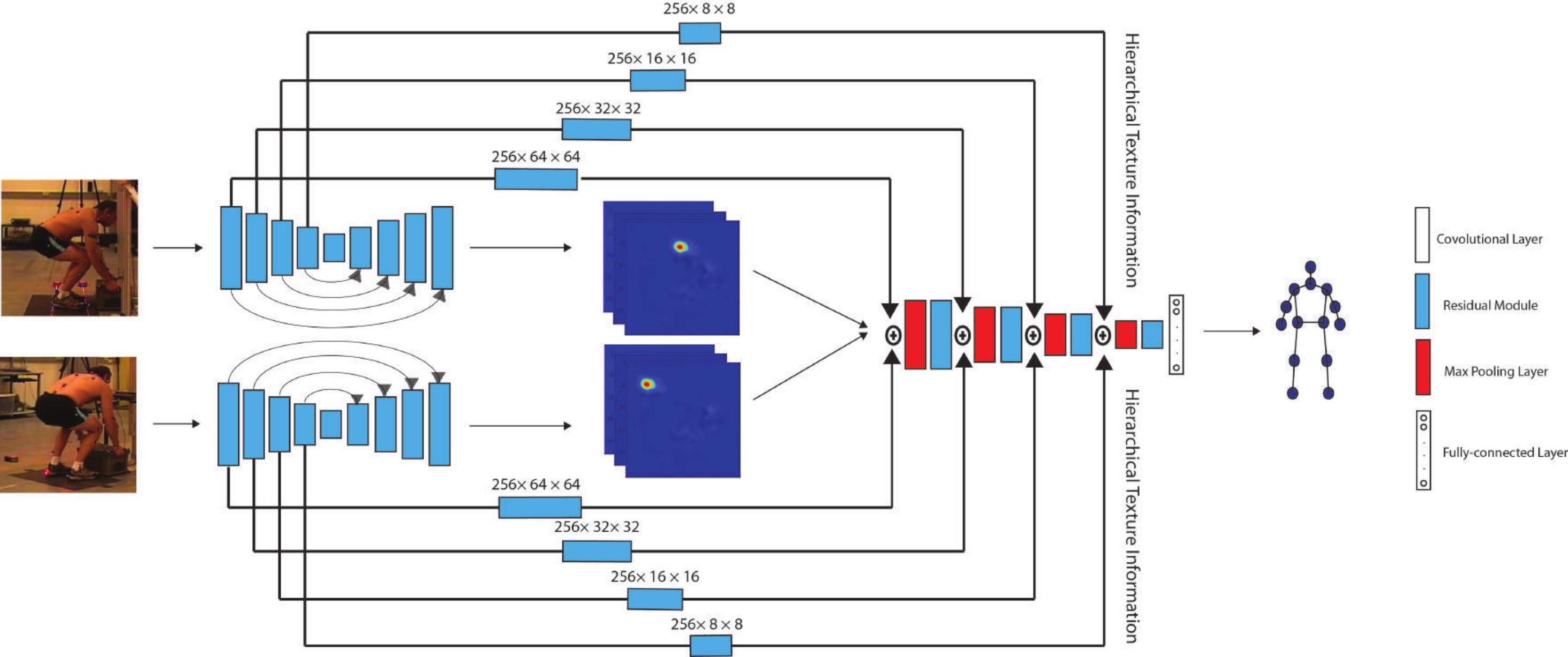}
      \caption{An overview of our approach. The "view-specific perceptron" network extracts both shape and hierarchical texture from different views; while the "multi-view integration" network synthesizes information from all available views to estimation 3D pose. Note that the hierarchical skip connections are not only shared locally inside the first network, but also shared globally between two networks for efficient and effective feature embedding.}
      \label{figurelabel}
   \end{figure*}
   
\section{METHODS}
In this work, we aim to predict the 3D body pose from multi-view RGB images. We proposed a deep learning based method for this purpose whose inputs are totally N different viewpoints around the subject and the output is the 3D coordination of the body joints, which define the pose. Fig. 2 shows an illustration of our approach for N=2, which consists of two networks: a "view-specific perceptron" network and a "multi-view integration" network. The first network extracts both shape (2D pose) and hierarchical texture information independently from each view, while the second network synthesizes these information from all available views to infer the 3D pose. 
                                  
\subsection{View-specific Perceptron Network}
View-specific perceptron network extracts rich information independently from each view, which includes not only 2D shape but also hierarchical texture information for 3D inference in the next step. Each 2D body pose is represented by J heatmaps, where J is the number of body joints. Let $x^i \in \mathbb{R}^{W \times H \times 3}$ be the input RGB image for view i, $t_s^i \in \mathbb{R}^{W_s \times H_s \times L_s}(s=1,...,S)$  be s-th texture feature map for view i, and $h_j^i \in \mathbb{R}^{W_h \times H_h \times L}(j=1,...,J)$ be j-th joint heatmap for view i. Then, view-specific perceptron network (f) for i-th view is a mapping as follow:
$$((h_1^i,...,h_J^i ),(t_1^i,...,t_S^i ))=f(x^i).$$
The intermediate supervision is performed by pixel-wise heatmap loss:
$$\mathcal{L}_{2d}^i=1 \ / J \sum_{j=1}^{J} ||h_j^i- \widehat{h_j^i}||,$$
where $||.||$ is the Euclidean distance and $\widehat{h_j^i}$ is rendered from the ground truth 2D pose through a Gaussian kernel with mean equal to the ground truth and variance one.\\
We use Hourglass architecture \cite{newell2016stacked}, which has achieved state-of-the-art performance on large scale human pose datasets. Hourglass network \cite{newell2016stacked} comprises of encoder and decoder. Encoder processes the input image with convolution and pooling layers to generate low resolution feature maps and the decoder processes the low resolution feature maps with up-sampling and convolution layers to construct the high resolution heatmaps for each joint. One of the key components of the Hourglass network \cite{newell2016stacked} is the “skip connections”, the feature maps before each pooling layer, which are directly added to the counterpart in the decoder in order to prevent the loss of high resolution information in the encoder.  These hierarchical skip connections of the network share rich texture information in different scales. So, we propose to employ them for a more efficient 3D inference by feeding them to the multi-view integration network. We will show soon that they allow for a richer gradient signal and can provide more 3D cues compare to using only heatmaps or a combination of heatmaps and unprocessed input images.
\subsection{Multi-view Integration Network}
The multi-view integration network integrates information from multiple views to synthesize 3D pose estimation. The input of this network is the concatenation of the outputs of the view-specific perceptron network for N different views and the output is the 3D pose. Each 3D pose skeleton $p \in \mathbb{R}^{3 \times J}$  is defined as a set of joints coordination in 3D space. So multi-view integration network (g) is a mapping as follow:
\begin{multline*}
(\hat{p})=g(concat(h_1^1...…,h_1^N )...…,concat(h_J^1...…,h_J^N)\\
,concat(t_1^1,.........,t_1^N ),......,concat(t_S^1,...,t_S^N )).
\end{multline*}

\noindent By assuming that 3D joints annotations are available for training dataset, the loss function can be defined as
$$\mathcal{L}_{3d}^i=1 \ / J \sum_{j=1}^{J} ||p_j^i- \widehat{p_j^i}||,$$
where $p_j$ and $\hat{p}_j$ are ground truth and estimated 3D coordinate of joint j, respectively.\\

\noindent We propose a bottom up data driven method that directly generates the 3D pose skeleton from the outputs of the view-specific perceptron network. The multi-view integration network is designed as an encoder. We tested two types of encoders: first, an encoder consists of a series of convolutional layers with kernel and stride size of 2 in which the resolution of the feature maps are half at each layer; second, an encoder similar to the first part of the Hourglass network \cite{newell2016stacked}, which includes max-pooling layers and standard convolutional layers are replaced by a stack of residual learning modules \cite{he2016deep}. In the rest of this paper, we call the first and second network architectures as “simple encoder” and “half-hourglass” network, respectively. For both network architectures, the encoder output is then forwarded to a fully-connected layer with output size of $3 \times J$ for estimating 3D pose skeleton and measuring the loss function for training. Fig. 3 shows the schematic comparison of simple encoder and half-hourglass architecture in a simplified setting. It will be shown that, half-hourglass network that benefits from residual modules and periodically insert of max-pooling layer can provide more accurate 3D pose compare than the simple encoder network.

\subsection{Hierarchical Skip Connections}
Inferring a 3D pose from joints heatmap as the only intermediate supervision, which is a widely used strategy in previous studies \cite{pavlakos2017coarse, tome2017lifting}, is inherently ambiguous. This ambiguity comes from the fact that usually exist multiple 3D poses corresponded to a single 2D pose. In order to overcome this challenge in 3D pose estimation, joints heatmaps can be combined with either input image or its lower-layer features \cite{tekin2017learning, he2016deep, zhou2017towards} as the intermediate supervision. While taking the input image into account can provide more information compare to only joints heatmap, combining hierarchical texture information, learnt at the view-specific perceptron network, extract additional cues \cite{zhou2017towards}. As a result, we propose to leverage skip connections of the Hourglass network \cite{newell2016stacked} to multi-view integration network. In our proposed framework, each of the four skip connections produced in the encoder part of the Hourglass network \cite{newell2016stacked}, is processed with residual modules and summed with the counterpart in the half-hourglass network (fig. 2). In order to handle multi-view setup, each skip connection should be concatenated across the views before being provided as inputs for the network. 

   \begin{figure}[t]
      \centering
      \includegraphics[width=2.5in]{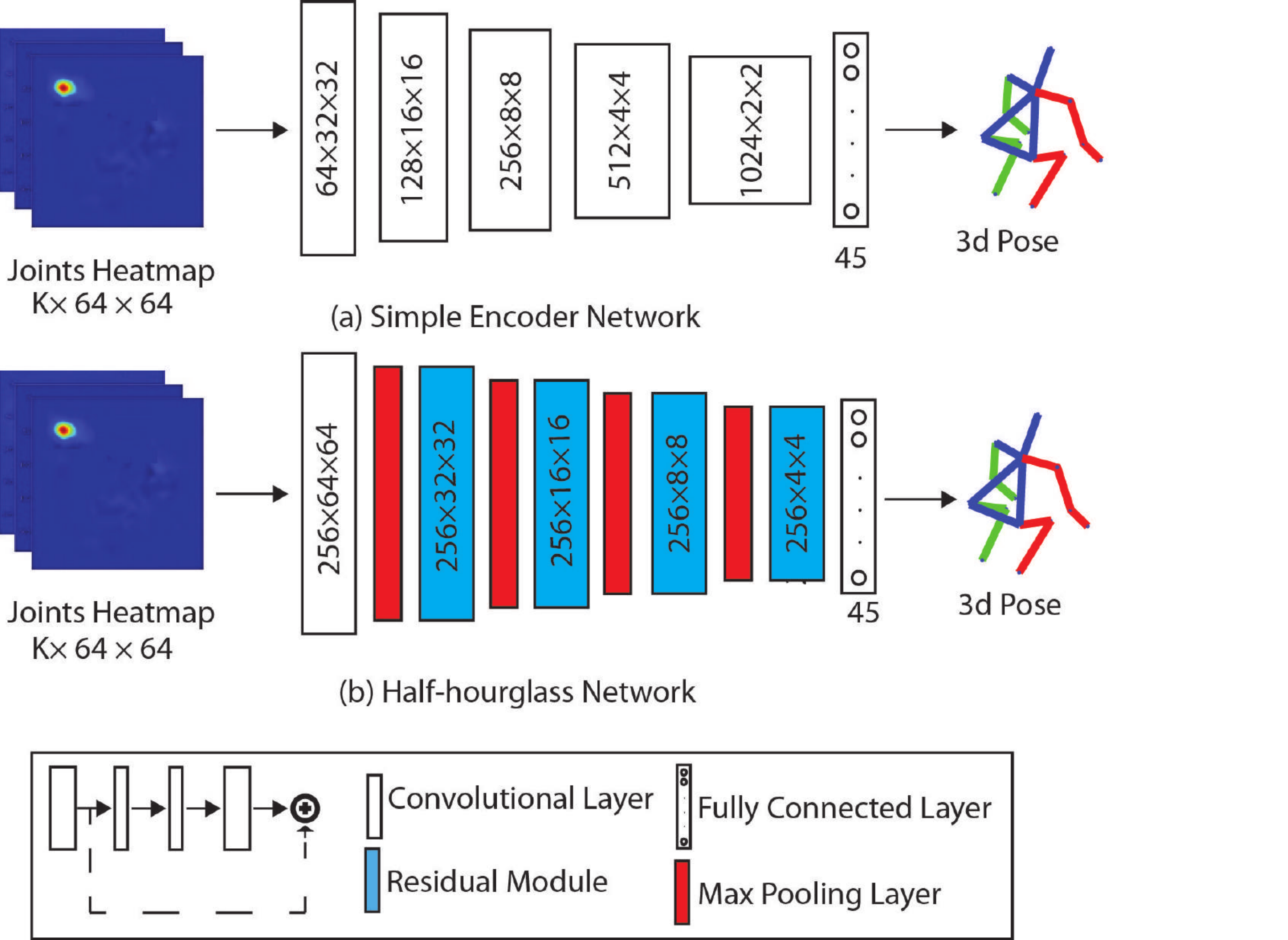}
      \caption{Architecture comparison of (a) simple encoder and (b) half-hourglass design for multi-view integration network. The numbers inside each layer illustrate the corresponding size of the feature maps (number of channels × resolution) for convolutional layers and residual modules [36] and the number of neurons for fully connected layers. The architecture on the bottom represents residual module that is used throughout the network.}
      \label{figurelabel}
   \end{figure}

%%%%%%%%%%%%%%%%%%%%%%%%%%%%%%%%%%%%%%%%%%%%%%%%%%%%%%%%%%%%%%%%%%%%%%%%%%%%%%%%
\section{MATERIALS}

\subsection{Participants and Data Acquisition}

Our lifting dataset consists of 12 subjects. Each subject performed various symmetric and asymmetrical lifting tasks in a laboratory at a self-selected speed. A motion tracking system was used to capture 3D coordination of body joints. Two digital camcorder (GR-850U, JVC, Japan) with resolution $720 \times 480$ pixel, synchronized with the motion tracking system also recorded the tasks from 90 degree (side view) and 135 degree view positions. Subjects lifted a plastic crate ($39 \times 31 \times 22$ cm) weighing 10 kg and placed it on a shelf without moving their feet. They performed three vertical lifting ranging from floor to knuckle height (FK), knuckle to shoulder height (KS) and floor to shoulder height (FS). Each vertical lifting range was combined with three end-of-lift angles (0, 30 and 60 degree), which is defined as the angle of the end position relative to the starting position of the box. For each combination of the lifting task, two repetitions were performed, providing a total of 18 lifts ($3 \times 3 \times 2$).

\subsection{Data Pre-processing}
To prepare the training images, we follow \cite{peng2015piefa} to down sample images from videos. Each video includes 200 frames with 30 fps rate, where only odd frames are employed in this study to prevent overfitting. All of the images are adjusted to $256 \times 256$ pixels and are cropped such that the subject is located at the center.\\
3D joints annotation are provided by a motion capture system. We selected 23 markers to define 14 joints including head, neck, left/right shoulder, left/right elbow, left/right wrist, left/right hip, left/right knee, and left/right ankle and only used the trajectory of these joints for training the network. The coordination of each joint is normalized from zero to one over the whole dataset. After pre-processing, the data structure consists of the cropped images and corresponding normalized 3D joints annotation for every odd frame of the videos.

\subsection{Training Strategy}

We propose a two-stage training strategy that we found more effective instead of an end-to-end training for the whole network from the scratch. At the first stage, we used the Hourglass model \cite{newell2016stacked} for MPII \cite{andriluka20142d} and fine-tuned it on our lifting dataset with learning rate of 0.00025 for five epochs. At the second stage, multi-view integration model was trained from scratch on our lifting dataset by using two-view images and corresponding normalized 3D pose skeleton. The models were trained with learning rate of 0.0005 for 50 epochs. \\
In order to evaluate the performance of the network for both single-view and two-view setups, we ran two experiments: first, the network was trained for a single-view setup using both 90 and 135 degree view separately; second, the network was trained for a two-view setup utilizing both views together as inputs for the network. In all of the experiments, repetition one of all the subjects and lifting tasks were used as training dataset and repetition two as testing dataset. It will be shown that the proposed network is robust and can achieve high performance in both experiments.

\subsection{Evaluation Protocol}
Following the evaluation protocol of the publicly available datasets \cite{sigal2010humaneva} we calculate 3D human pose estimation error based on the average Euclidean distance between estimated 3D joints coordination and corresponding ground-truth data obtained from a motion capture system.

%%%%%%%%%%%%%%%%%%%%%%%%%%%%%%%%%%%%%%%%%%%%%%%%%%%%%%%%%%%%%%%%%%%%%%%%%%%%%%%%
\section{RESULTS}
In this chapter we present experimental results on our lifting dataset. We used Pytorch interface in this work and training and testing have been performed on a machine with NVIDIA Tesla K40c and 12 GB RAM. We executed three experiments to study the effect of three different factors on the accuracy of results. First, three variants of multi-view integration inputs; including joint heatmaps, joints heatmaps plus input images, and joints heatmaps plus skip connections, are tested to assess how the accuracy changes by feeding more 3D cues to this network. Second, we tested two network structures for multi-view integration, namely simple encoder and half-hourglass, to evaluate the influence of using max-pooling and residual learning modules instead of standard convolutional layers on our dataset. Third, single-view and two-views training are performed to study the robustness of the method as a function of the number of cameras. Finally, we chose the method with the highest performance (fig. 2) and applied it on HumanEva-I dataset \cite{sigal2010humaneva} and compared our results with other state-of-the-art method on this dataset.

\subsection{Different Variants of Multi-view Integration Inputs}

   \begin{figure}[t]
      \centering
      \includegraphics[width=3in]{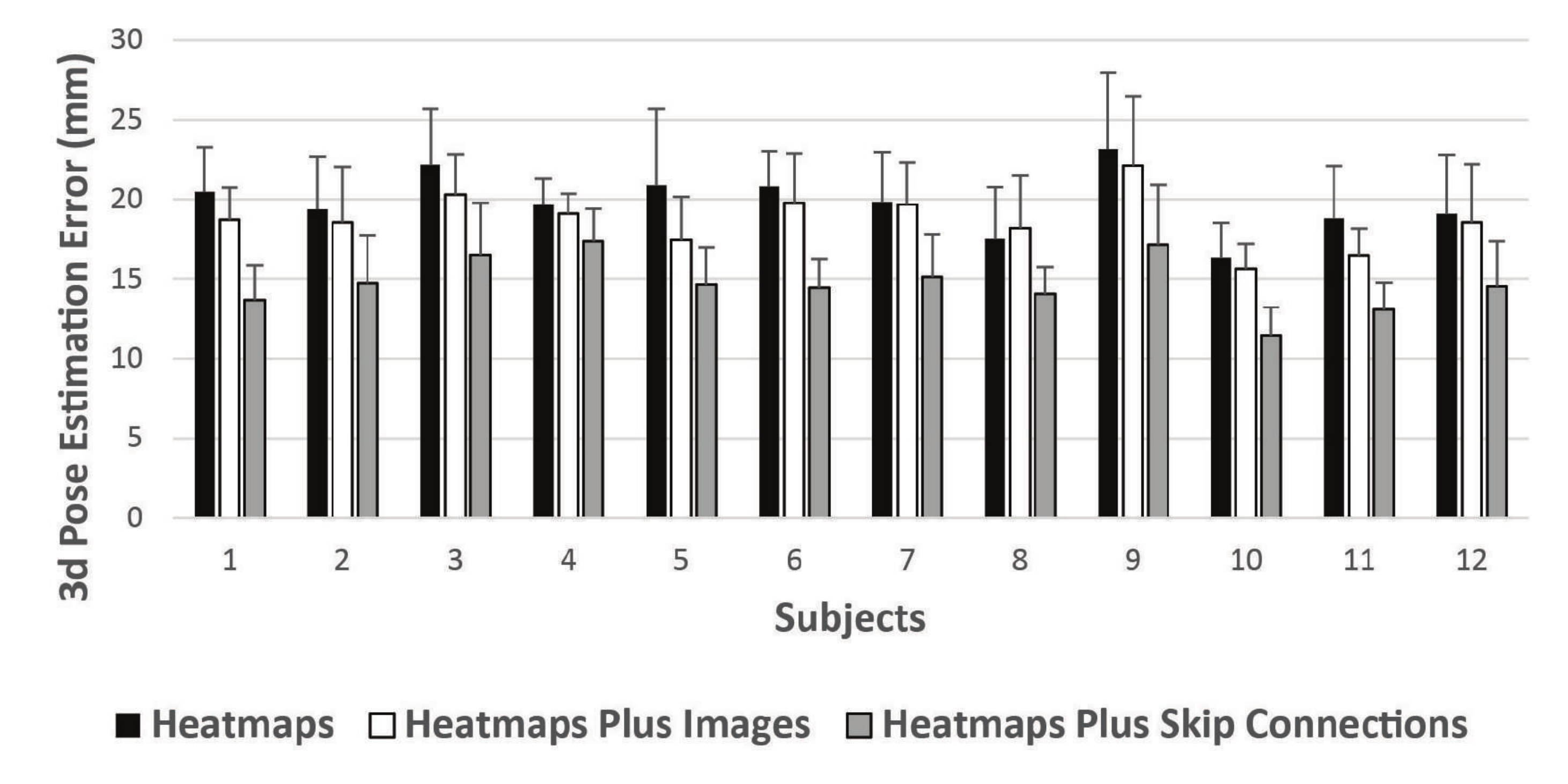}
      \caption{Average of the 3D pose estimation error of different subjects for three variants of multi-view integration inputs. Bars show the variance.}
      \label{figurelabel}
   \end{figure}

Fig. 4 illustrates the 3D pose estimation error of different subjects for three combinations of input for multi-view integration network. It can be seen that summing up skip connections with feature maps in-between residual modules can achieve the highest accuracy. The error reduction of combining input images with joints heatmaps is only \%6 ($19.82 \pm 3.77$ mm vs $18.69 \pm 3.25$ mm), compare to \%26 ($19.82 \pm 3.77$ mm vs $14.72 \pm 2.96$ mm) error reduction by combining skip connections and joint heatmaps as input to the multi-view integration network. While input images might provide noisy information for the network, these skip connection features can extract semantic information at multiple levels of 2D pose estimation and provide more cues for 3D pose inference. 

\subsection{Simple Encoder vs Half-hourglass}

Fig. 5 illustrates the 3D pose estimation error of different subjects for simple encoder and half-hourglass architectures. The average error over the whole dataset is $25.98 \pm 6.39$ mm and $19.82 \pm 3.77$ mm for these networks, respectively. We found that using half-hourglass network that benefits from residual modules and periodically insert of max-pooling layer reduces the error by \%24. This happens due to the fact that networks with residual modules gain accuracy from greatly increased depth and addressing the degradation problem \cite{he2016deep}. In addition, inserting max-pooling layer in-between successive convolutional layers reduces the number of parameters and computation in the network, and control overfitting. For qualitative results, we have provided representative 3D poses predicted by our proposed method (half-hourglass architecture and using both heatmaps and skip connections as the inputs for multi-view integration network) in figure 6. It can be seen that even for posture with self-occlusion, our method is able to predict the pose accurately.

   \begin{figure}[t]
      \centering
      \includegraphics[width=3in]{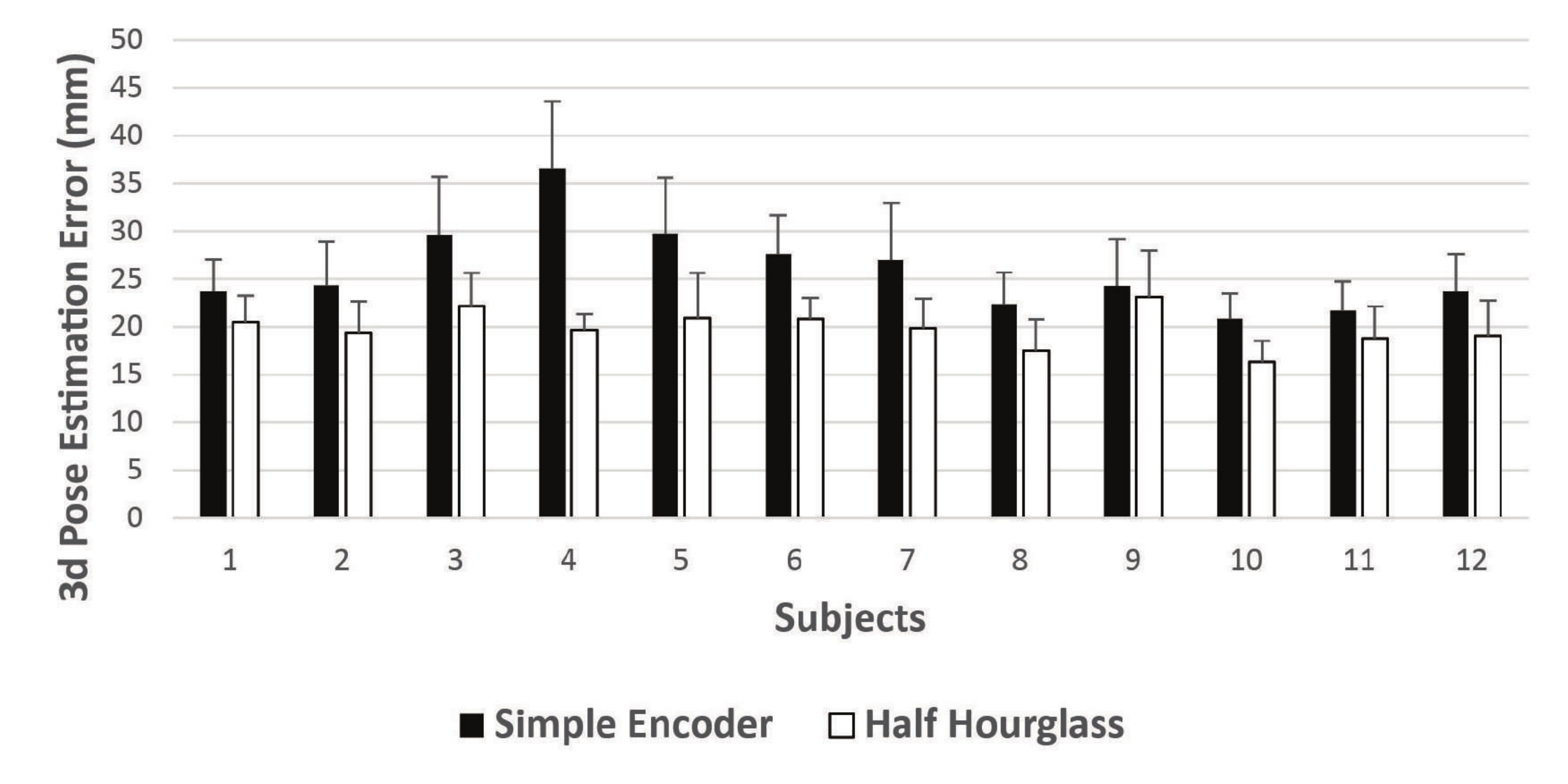}
      \caption{Average of the 3D pose estimation error of different subjects for simple encoder and half-hourglass architecture. Bars show the variance.}
      \label{figurelabel}
   \end{figure}
   
   \begin{figure*}[t]
      \centering
      \includegraphics[width=5in]{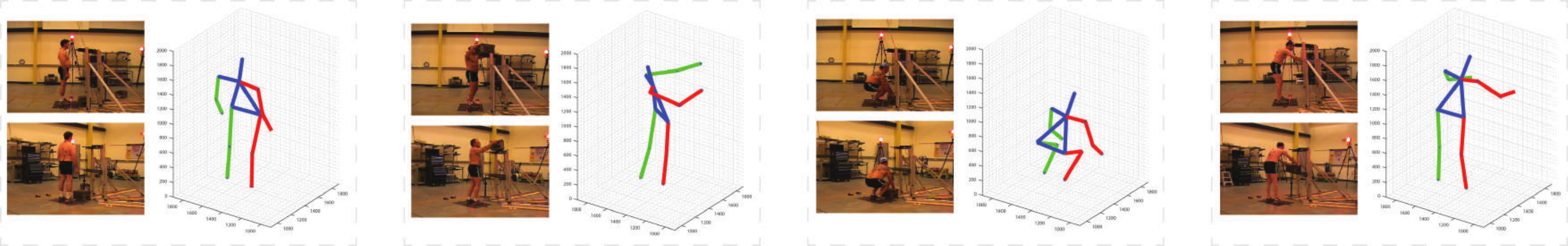}
      \caption{Qualitative results for lifting datasets. Each dashed box represents a scenario: \textbf{Left-} multi-view images, \textbf{Right-} corresponding estimated 3D pose.}
      \label{figurelabel}
   \end{figure*}

\subsection{Single-view vs Two-view Training}
Fig. 7 illustrates the 3D pose estimation error of different subjects for single-view and two-view setups. For single-view setup, the network is trained and tested on both views. As was expected, we found that feeding two-view images as inputs to the network increases the accuracy. Since self-occlusion is a main source of ambiguity in pose estimation and can be addressed by using multiple cameras. For biomechanical applications due to the need of very accurate estimated pose, using multiple cameras is crucial \cite{mundermann2006evolution}.

   \begin{figure}[t]
      \centering
      \includegraphics[width=3in]{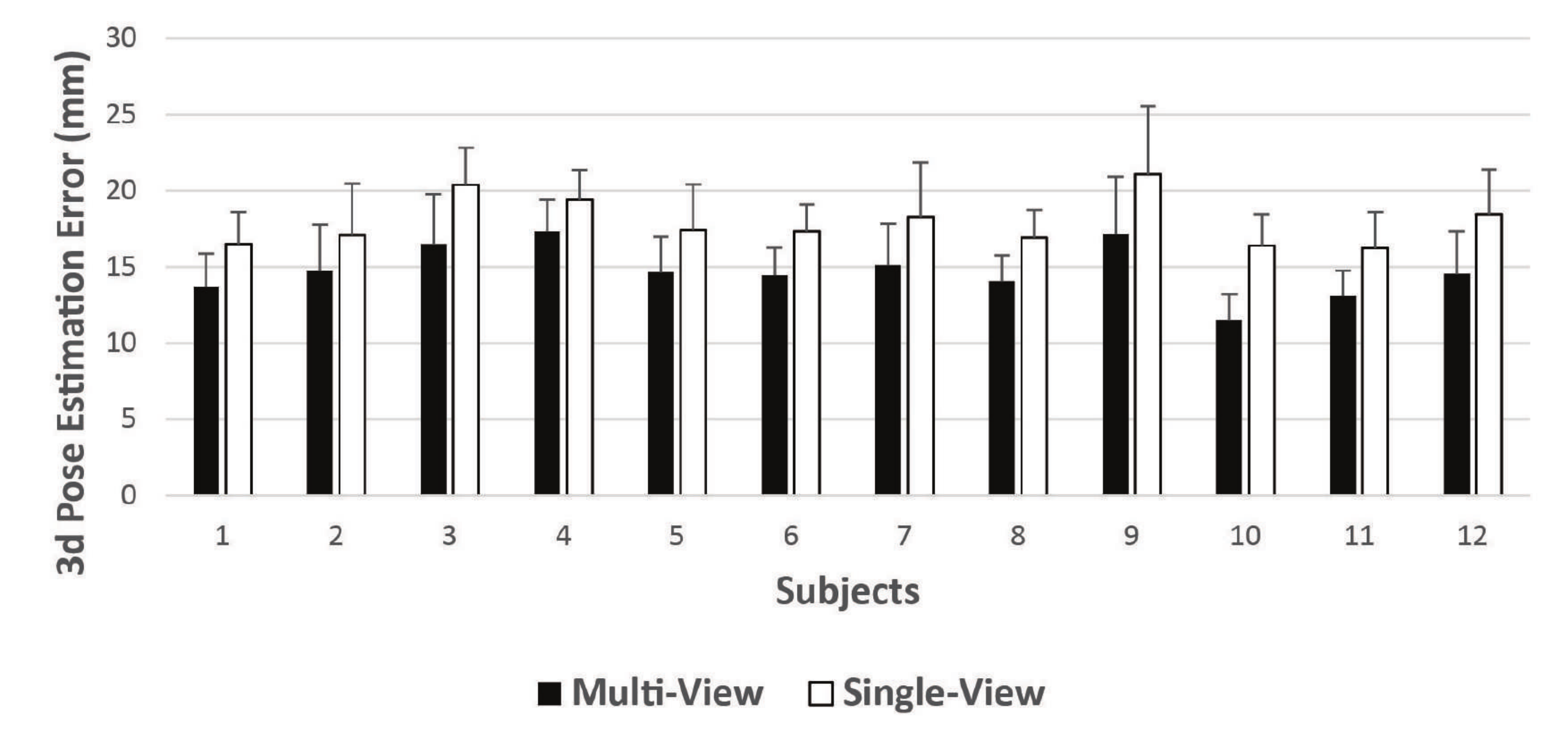}
      \caption{Average of the 3D pose estimation error of different subjects for single-view and multi-view setups. Bars show the variance. }
      \label{figurelabel}
   \end{figure}

\subsection{Compare with Other Work}
 We applied our method on HumanEva-I dataset \cite{sigal2010humaneva} to be able to compare it with state-of-the-art multi-view methods. We followed the standard protocol of the dataset for evaluation and compared the performance of our method on walking and jogging sequences of subjects 1 to 3. As can be seen from the results in table 1, our proposed deep convolutional neural networks obtain the best result for both walking and jogging sequences.

%%%%%%%%%%%%%%%%%%%%%%%%%%%%%%%%%%%%%%%%%%%%%%%%%%%%%%%%%%%%%%%%%%%%%%%%%%%%%
\section{CONCLUSION}
In this work, we presented a multi-view DNN-based method for 3D pose estimation. In part of this study, we introduced an approach to integrate hierarchical texture information with estimated joints heatmap to infer 3D pose. We tested several different network architectures to analyze the influence of various parameters on the accuracy of the results. With optimal network architecture, which consists of half-hourglass architecture for multi-view integration network combined with skip connections, we estimated the pose on our lifting dataset with $14.72 \pm 2.96$ mm error compared to marker-based motion capture system. The results on a publicly available dataset (HumanEva-I \cite{sigal2010humaneva}) also shows the superior performance of our approach. This result demonstrates the applicability of deep learning techniques in the context of biomechanical analysis. For future work, we want to use the estimated 3D pose for further biomechanical analysis like calculating joints force and moment in order to automatically detect not-safe lifting in the workplaces with the aim of preventing injuries.

\begin{table}[h]
\caption{COMPARISON OF OUR METHOD WITH STATE-OF-THE-ART METHODS ON HUMANEVA DATASET \cite{sigal2010humaneva}. ‘NA’ INDICATES THAT RESULTS ARE NOT REPORTED FOR THE CORRESPONDING ACTION IN THE ORIGINAL PAPER.}
\label{table_example}
\begin{center}
\begin{tabular}{|c|c|c|c|c|c|c|}
\hline
    \multirow{2}{*}{Methods} & \multicolumn{3}{c|}{Walking} & \multicolumn{3}{c|}{Jogging}\\ \cline{2-7}
     & S1 & S2 & S3 & S1 & S2 & S3 \\
   % \hline
\hline
Elhayak et. al. \cite{elhayek2015efficient} & 66.5 & NA & NA & NA & NA & NA\\
Amin et. al. \cite{amin2013multi} & 54.5 & 50.2 & 54.7 & NA & NA & NA\\
Sedai et. al. \cite{sedai2013gaussian} & 42.4 & 34.1 &62.9 & 70.9 & 50.6 & 55.1 \\
Zhang et. al. \cite{zhang2014robust} & 44.3 & 58.4 & 66.0 & 55.4 & 68.2 & 57.5\\
Tekin et. al. \cite{tekin2016direct} & 37.5 &25.1 & 49.2 & NA & NA & NA\\
\textbf{Ours} & \textbf{40.4} & \textbf{23.5} & \textbf{33.4} & \textbf{43.0} & \textbf{45.1} & \textbf{30.9}\\
\hline
\end{tabular}
\end{center}
\end{table}

%%%%%%%%%%%%%%%%%%%%%%%%%%%%%%%%%%%%%%%%%%%%%%%%%%%%%%%%%%%%%%%%%%%%%%%%%%%%%%%%
\section{ACKNOWLEDGMENTS}

This work was supported in part by NSF (CMMI 1334389, IIS 1451292, IIS 1555408, and IIS 1703883). The lifting data collection was conducted at Liberty Mutual Research Institute for Safety.

%%%%%%%%%%%%%%%%%%%%%%%%%%%%%%%%%%%%%%%%%%%%%%%%%%%%%%%%%%%%%%%%%%%%%%%%%%%%%%%%
\bibliographystyle{unsrt}
\bibliography{egbib}
%\printbibliography{}

\end{document}